\documentclass[twocolumn]{el-author}

\usepackage{epstopdf}

\newcommand{\ua}{\uparrow}
\newcommand{\nc}{\newcommand}
\nc{\da}{\downarrow} \nc{\hc}{\hat{c}} \nc{\hS}{\hat{S}}
\nc{\bra}{\langle} \nc{\ket}{\rangle} \nc{\eq}{equation (\ref}
\nc{\h}{\hat} \nc{\hT}{\h{T}}\nc{\be}{\begin{eqnarray}}
\nc{\ee}{\end{eqnarray}}\nc{\rd}{\textrm{d}}\nc{\e}{eqnarray}\nc{\hR}{\hat{R}}\nc{\Tr}{\mathrm{Tr}}
\nc{\tS}{\tilde{S}}\nc{\tr}{\mathrm{tr}}\nc{\8}{\infty}\nc{\lgs}{\bra\ua,\phi|}\nc{\rgs}{|\ua,\phi\ket}
\nc{\hU}{\hat{U}}\nc{\lfs}{\bra\phi|}\nc{\rfs}{|\phi\ket}\nc{\hZ}{\hat{Z}}\nc{\hd}{\hat{d}}\nc{\mD}{\mathcal{D}}
\nc{\bd}{\bar{d}}\nc{\bc}{\bar{c}}\nc{\mc}{\mathcal}\nc{\ea}{eqnarray}\nc{\mG}{\mathcal{G}}\nc{\bce}{\begin{center}}
\nc{\ece}{\end{center}}

\begin{document}

\title{Detection of Salient Regions in Crowded Scenes}
\author{M. K. Lim, C. S. Chan, D. Monekosso, P. Remagnino}


\abstract{The increasing number of cameras and a handful of human operators to monitor the video inputs from hundreds of cameras leave the system ill equipped to fulfil the task of detecting anomalies. Thus, there is a dire need to automatically detect regions that require immediate attention for a more effective and proactive surveillance. We propose a framework that utilises the temporal variations in the flow field of a crowd scene to automatically detect salient regions, while eliminating the need to have prior knowledge of the scene or training. We deem the flow fields to be a dynamic system and adopt the stability theory of dynamical systems, to determine the motion dynamics within a given area. In the context of this work, salient regions refer to areas with high motion dynamics, where points in a particular region are unstable.  Experimental results on public, crowd scenes have shown the effectiveness of the proposed method in detecting salient regions which correspond to unstable flow, occlusions, bottlenecks, entries and exits.}

\maketitle


\section{Introduction}
\label{sec:Intro}

\begin{figure} [h]
\begin{center}  
  \includegraphics[width=0.45\textwidth]{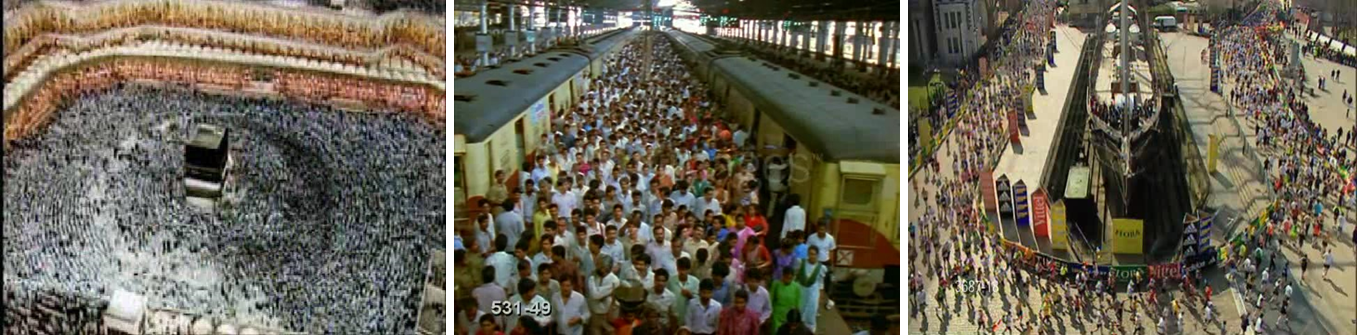}
\caption{Sample shots of the different scenarios of crowded scenes (pilgrimage, train station and marathon).} 
\label{fig:SampleCrowd} 
\end{center}
\end{figure}

Conventional CCTV monitoring by human operators becomes increasingly demanding as the average number of cameras deployed grows. Research findings have shown that besides fatigue and boredom, human attention tends to decline after 20 minutes. Therefore a high percentage of questionable activities, are often overlooked. This is made even more challenging when monitoring crowded scenes such as the footage of pilgrimage as shown in Fig.\ref{fig:SampleCrowd}. Anomalous activity or behavior in a crowded scene can be very subtle and imperceptible to a human operator \cite{Challenger09}. Thus, an automated detection of suspicious regions is critical to direct the attention of security personnel to areas that require further investigation. Automated saliency detection is useful in numerous applications, such as identifying bottlenecks, which may help in avoiding congestion or evacuation planning.

Most work in saliency detection are focused on detection of salient regions in an image, where saliency originates from visual uniqueness and is often deciphered from image attributes such as colour, gradient and edges \cite{Xu05}. Saliency in image differs from saliency in video sequence and using image attributes alone is not sufficient to infer the motion dynamics of crowded scenes. Boiman \emph{et.al} in \cite{Boiman07} proposed a graphical inference algorithm to detect irregularities in videos based on the spatio-temporal features of different scales of image patches. While their method works well in detecting irregularities in both image and videos, they are not utilising the benefit of motion information from videos and does not cope with large scale crowd scenes.   

Research into the motion dynamics in dense crowds, \cite{saad07,zhou12, rodriguez11} is limited to learning the coherent motion patterns or dominant crowd flows, where regions with similar motion information are grouped into the same cluster. In contrast, our method ignores dominant flows and instead, is focused on regions with high motion dynamics or unstable, to infer salient regions. The closest work to ours, thus far, is by Loy and \emph et. al. in \cite{Loy12} where dominant flows are suppressed, while focusing on motion flows that deviate from the norm. However, their method which is based on spectral analysis of motion flows is only reliable when detecting obvious saliency such as crowd instability and counter flow detection. They do not deal with more subtle scenarios of saliency such as bottlenecks and occlusions. In \cite{solmaz12}, a set of rules is applied to the eigenvalue map to discover the different motion behaviors such as bottleneck and arch. While their method is able to discriminate the different types of saliency, it is restricted by pre-defined conditions and requires characteristic flows.  Our method on the other hand is not restricted by the set of rules, and assumes anomaly when a particular region exhibits high motion dynamics.     

This work extend the definition of salient regions to include subtle anomaly which corresponds to bottlenecks and occlusions. In addition, we introduce the simple, yet effective idea of amplifying regions with unstable motion instead of disregarding them as noise. This alludes to the social behaviour of humans in crowds. In a dense crowded scene, the motions of individuals tend to follow the regular or dominant flow of a particular region due to the physical constraints of the environment (i.e. path, junction) and the social conventions of crowd dynamics. We can therefore consider the possibility of irregularities or anomalies occuring in the scene, when the motion dynamics of individuals differ from its close neighbours. In our work, we first magnify and then examine the unstable regions by performing a two stages segmentation process to infer salient regions. Our method does not rely on tracking each object or on prior learning, thus can adapt to the environment over time.


\section{Magnification of Unstable Flow}
\label{sec:Solution}

We firstly estimate the velocity field at each point, $V(p) = (u,v)$ by employing the dense optical flow algorithm in \cite{liu08}. The velocity components of each point are accumulated and an average velocity is calculated within an interval of time, comprising $\tau$ frames.

\begin{equation}\label{eq:mean01}
\overline{V} = \{\overline{u},\overline{v}\} = \{\frac{1}{\tau}\sum _t^{t+\tau} u_p, \frac{1}{\tau}\sum _t^{t+\tau} v_p\}
\end{equation}

While the mean velocity field may be a good indicator of the global flow of individuals in a crowd, it is unstructured and may change over time. A particle advection process is implemented to keep track of the velocity changes for each point, $p$ along its velocity field, $(u,v)$.

\begin{equation}\label{eq:ode03}
\frac{d\vec{x}_p}{dt} = u_p(x_p,t; t_0,x_0) 
\end{equation}
\begin{equation}\label{eq:ode05}
\frac{d\vec{y}_p}{dt} = v_p(y_p,t;t_0,x_0) 
\end{equation}
subject to
\begin{equation}\label{eq:ode02}
p=p_0 \;\;\; at \;\;\; t=t_0
\end{equation}

The suffix, $p$ indicates the motion of a particular particle or point, $p$. Assuming that the initial position of $p_0$ is the mean velocity fields, $(\overline{u}, \overline{v})$, we deem the dynamic system as an initial value problem. Thus, the pathlines which trace the points from their $x_0$ and $y_0$ positions at time, $t_0$ to their positions, $x_t$ and $y_t$ at time, $t$ can be solved using the fourth-order Runge Kutta Scheme (RK4) as in \cite{kennedy03}.

We adopted the Jacobian method as in \cite{haller00}, to measure the separation between particle's pathlines which are seeded spatially close to a point, $p$, within a time instance, $\tau$. Assuming that a particle's position is slightly shifted from $p$ at time $t_0$, to $p+\Delta{p}$ at time $t+\tau$, the Jacobian, denoted as $\nabla{ F_t (p)}$, multiplied by the offset, $\Delta{p}$, indicates the coordinate offset at time $t+\tau$. This is based on the assumption that the displacement, $\Delta{p}$, is small. The Jacobian of the flow map is computed by the partial derivatives of $d\vec{x}$ and $d\vec{y}$, where:

\begin{equation}\label{eq:jacobian01}
\nabla{F_t(p)} = \begin{bmatrix}
\frac {\partial{d\vec{x}}}{\partial{x}} & \frac {\partial{d\vec{x}}}{\partial{y}} \\
\frac {\partial{d\vec{y}}}{\partial{x}} & \frac {\partial{d\vec{y}}}{\partial{y}}
\end{bmatrix} 
\end{equation}

According to the theory of linear stability analysis, the square root of the largest eigenvalue, $\lambda_t(p)$ of $F_t(p)^T F_t(p)$ indicates the maximum offset or displacement if the particle's seeding location is shifted by one unit as it satisfies the condition that $ln \lambda_t(p) > 0$. In the context of this study, a large eigenvalue indicates that the query point is unstable, and vice versa for a small eigenvalue. Since we are interested in regions that have high motion dynamics, based on the maximum eigenvalue, we can compute the stability of a point within its spatially close neighbouring points using equation:
\begin{equation}\label{eq:stability01}
\phi_t = \frac{1}{\mid \tau \mid} log \sqrt{\lambda_t(p)}
\end{equation}

We propose two stages of segmentation that combine the output of fine and coarse segmentation obtained from the local and global flow segmentation steps, followed by a flow magnification of regions with high motion instability to synthesize the signal, where $\beta$ is the magnification factor and $\alpha$ is the segmentation threshold:

\begin{equation}\label{eq:stability02}
\hat{\phi}_t= 
\begin{cases}
    \beta . \phi_t,& \text{if } \phi_t\geq \alpha\\
    (1-\beta) . \phi_t,              & \text{otherwise}
\end{cases}
\end{equation}


\section{Experiment: Instability Detection}
\label{sec:Experiment}

\begin{figure}[ht]
\centering
\subfigure[In addition to the ground truth unstable region (as enclosed in red and yellow bounding boxes), our method detected salient regions caused by bottlenecks (as highlighted in red blobs).]{
    \includegraphics[width=0.48\textwidth]{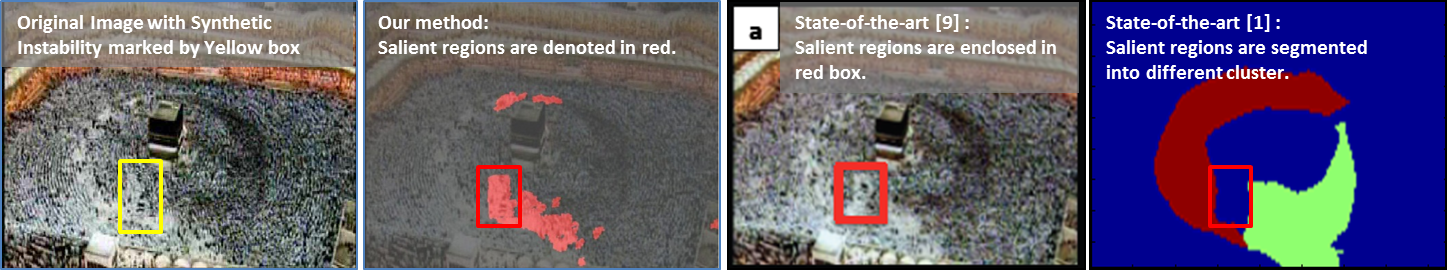} 
    \label{fig:subfigresult1}
}
\subfigure[Our method detects salient regions that may be caused by sudden slow down or potential danger due to high densities and instability.]{
    \includegraphics[width=0.48\textwidth]{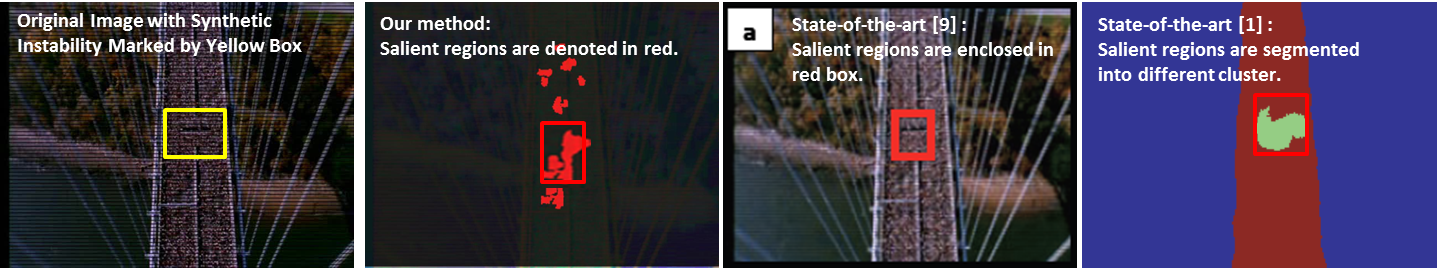} 
    \label{fig:subfigresult1}
}
\caption{Sample comparison results (with synthetic noise).}
\label{fig:subfigresult}
\end{figure}

A set of 4 test sequences which comprises large scale crowd scenes were used for evaluation. The first sequence is obtained from the National Geographic documentary, 'Inside Mecca', while the second depicts a marathon scene. Synthetic noise was injected into both scenes to simulate instability in the motion of the crowd.  A comparison between our work, Loy \emph{et al.} \cite{Loy12} and Ali \emph{et al.} \cite{saad07} is performed. It is observed that all three methods are able to detect instability successfully as indicated by the red bounding boxes in Fig. \ref{fig:subfigresult}. However, our method identified additional regions as salient. After a thorough investigation of the original sequence by 3 operators, we noticed that these regions correspond to areas where there are strong interactions motion dynamics within the crowd. It is worth noting that manual annotation of ground truth salient region due to bottlenecks or turbulence is an open issue because these types of salient regions are considered subjective. In the pilgrimage sequence, we noticed that the additional salient regions detected by our method in fact do correspond to regions where there are strong interactions and motion dynamics. Due to the structure of the scene, or physical constraints of the Kaaba which is situated at the centre of the scene, the crowd tend to slow down their pace during the turning. In addition, the salient region detected near the synthetic instability is caused by the high motion dynamics near the entry and exit point. Thus, we argue that it is unfair to deem these detections as false positive. Instead, we presuppose if the detected regions can aid us in investigating and understanding the non-obvious motion dynamics of a scene.

\section{Bottleneck Detection}
\begin{figure}[ht]
\centering
\subfigure[Subtle saliency due to bottlenecks are detected by our method while state-of-the-art methods fail to detect these variations of saliency.]{
    \includegraphics[width=0.5\textwidth]{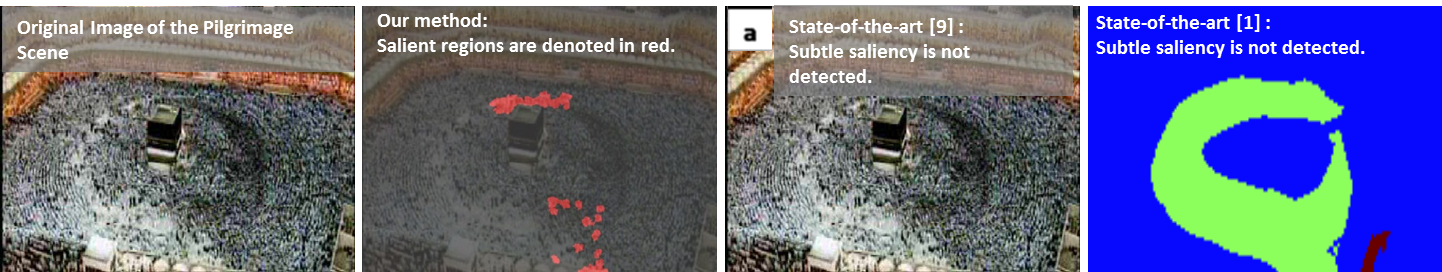} 
    \label{fig:subfigresultori1}
}
\subfigure[Subtle saliency due to high densities and stop-and-go waves. State-of-the-art methods fail to detect such saliency.]{
    \includegraphics[width=0.5\textwidth]{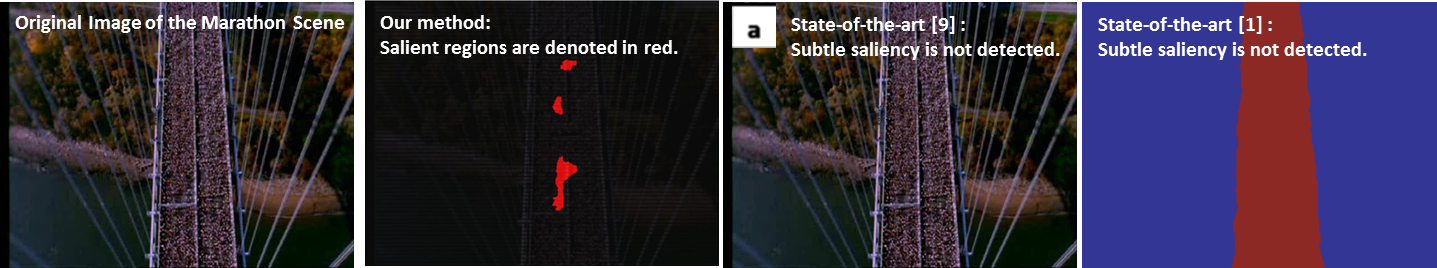} 
    \label{fig:subfigresultori2}
}
\caption{Sample comparison results (without synthetic noise).}.
\label{fig:subfigresultori}
\end{figure}

We further validated the capability of our method by using the original sequences, where no synthetic instability is introduced as shown in Fig. \ref{fig:subfigresultori}. The detections of bottleneck has tremendous potential as an indication of impending danger such as stampede taking place due the stop-and-go waves in the crowd motion.

\section{Occlusion and Turbulence Detection}

\begin{figure}[ht]
\centering
\subfigure[The detected regions grow across the frames as the motion dynamics of the school increases. This sequence comprises a school of fish manoeuvring towards the center of the scene.]{
    \includegraphics[width=0.45\textwidth]{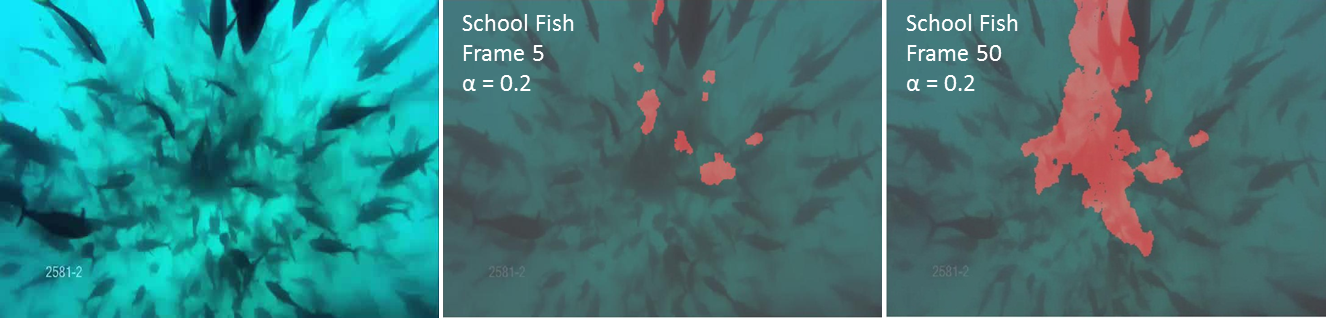} 
    \label{fig:subfigfish}
}
\subfigure[The street light which simulates the scenario of occlusion or barrier, is detected (in red).]{
        \includegraphics[width=0.45\textwidth]{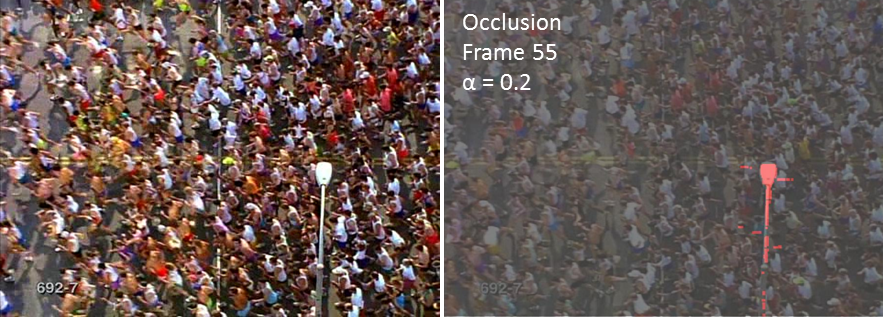} 
        \label{fig:subfiglight}
}
\caption{Qualitative results on other scenarios of saliency by our method.}
\label{fig:subfigconsolidated}
\end{figure}

We further test the robustness of the proposed method by using other scenarios of large scale crowd; the school of fish and marathon sequence (where there is a lamp post obstructing the flow); the results are as shown in Fig. \ref{fig:subfigconsolidated}.

\section{Conclusion}
We have proposed a framework that detects salient regions by observing the flow activities in a given scene with minimal observations. In addition, the proposed method eliminates the need to track each object individually or prior learning of the scene, which is critical for real-time operation. Experimental results show that the proposed method is not only able to detect salient regions that correspond to clear instability, but bottleneck and occlusion which is often difficult to be noticed by the naked eyes. The promising results obtained are definitely worthy of future investigation since it is able to detect regions that would otherwise go unnoticed by the human operator. The capability of the proposed method in spotting patterns of crowd activities that are subtle play a very important role in triggering real-time alarm to alert of potential danger such as stampedes, failed evacuations and crushes for operational decision making.

\vskip3pt
\ack{This work is supported by the University of Malaya Program Rakan Penyelidikan UM (PRPUM) under Grant CG065-2013.}

\vskip5pt
\noindent Mei Kuan Lim and Chee Seng Chan (\textit{Center of Image \& Signal Proc., Fac. of Comp. Sci. \& Info. Tech., Uni. of Malaya, MALAYSIA})
\vskip3pt
\noindent Dorothy Monekosso and Paolo Remagnino (\textit{Comp. and Info. Sys., Kingston University, UNITED KINGDOM})

\vskip3pt
\noindent E-mail: imeikuan@siswa.um.edu.my

\bibliographystyle{IEEEtrans}
\bibliography{ReferenceEL}

\end{document}